\documentclass{article}

 \usepackage[preprint]{neurips_2025}

\usepackage[utf8]{inputenc} 
\usepackage[T1]{fontenc}    
\usepackage{hyperref}       
\usepackage{url}            
\usepackage{booktabs}       
\usepackage{amsfonts}       
\usepackage{nicefrac}       
\usepackage{microtype}      
\usepackage{xcolor}         

\usepackage{setup/packages}
\usepackage{setup/custom_hyperref}
\usepackage{setup/custom}

\title{RewardUQ: A Unified Framework for Uncertainty-Aware Reward Models}

\author{%
    Daniel Yang$^1$\thanks{Equal contributions. Correspondence to \texttt{\{dayang,sstante,fredhardt,llibon\}@ethz.ch}.}
    \quad
    Samuel Stante$^1$\footnotemark[1]
    \quad
    Florian Redhardt$^1$\footnotemark[1]
    \quad
    Lena Libon$^1$\footnotemark[1]
    \\ \bfseries
    Parnian Kassraie$^1$
    \quad
    Ido Hakimi$^{1,2}$
    \quad
    Barna Pásztor$^{1,2}$
    \quad
    Andreas Krause$^{1,2}$
    \\\\
    $^1$ ETH Zurich
    \quad
    $^2$ ETH AI Center
}

\begin{document}


\maketitle
\suppressfloats

\begin{abstract}
Reward models are central to aligning large language models (LLMs) with human preferences.
Yet most approaches rely on pointwise reward estimates that overlook the epistemic uncertainty in reward models arising from limited human feedback.
Recent work suggests that quantifying this uncertainty can reduce the costs of human annotation via uncertainty-guided active learning and mitigate reward overoptimization in LLM post-training.
However, uncertainty-aware reward models have so far been adopted without thorough comparison, leaving them poorly understood.
This work introduces a unified framework, \emph{RewardUQ}, to systematically evaluate uncertainty quantification for reward models.
We compare common methods along standard metrics measuring accuracy and calibration, and we propose a new ranking strategy incorporating both dimensions for a simplified comparison.
Our experimental results suggest that model size and initialization have the most meaningful impact on performance, and most prior work could have benefited from alternative design choices.
To foster the development and evaluation of new methods and aid the deployment in downstream applications, we release our open-source framework as a Python package.
Our code is available at \url{https://github.com/lasgroup/rewarduq}.
\end{abstract}

\addtocontents{toc}{\protect\setcounter{tocdepth}{0}}

\section{Introduction}\label{sec:introduction}

\looseness=-1
Reinforcement learning from human feedback (RLHF) is a key component for aligning large language models (LLMs) with human preferences to ensure they are safe and helpful~\citep{ziegler2019fine, ouyang2022training, bai2022training}.
The standard RLHF process first trains a reward model on a dataset of pairwise comparisons to learn the underlying preferences, and then uses this model to align the LLM policy with reinforcement learning (RL) algorithms~\citep{christiano2017deep}.
However, the success of RLHF heavily relies on the quality of the reward model.
This poses several challenges, as collecting high-quality human preference data is expensive and reward models trained on limited and noisy datasets are imperfect~\citep{casper2023open}.
Additionally, aligning an LLM with such an imperfect reward model can lead to reward hacking, where the LLM overoptimizes flawed rewards rather than intended human preferences~\citep{eisenstein2024helping, coste2023reward, gao2023scaling, amodei2016concrete}. 

Uncertainty quantification (UQ) for reward models emerged as a promising way to address these issues by explicitly modeling epistemic uncertainty arising from limited preference data.
Recent work leverages uncertainty-aware reward models to mitigate reward hacking by penalizing~\citep{lou2025uncertaintyaware, zhai2024uncertaintypenalized, yan2024rewardrobust, houliston2024uncertaintypenalized, banerjee2024reliable, coste2023reward} or filtering~\citep{sun2024uncertainty, lou2025uncertaintyaware} uncertain samples.
Reward uncertainty estimates are also leveraged to improve sample efficiency and reduce data collection costs through active learning in the reward modeling \citep{melo2024deep, dwaracherla2024efficient, das2024active} or alignment step \citep{mehta2025sample, muldrew2024active, liu2024sampleefficient, gleave2022uncertainty, liang2021reward, christiano2017deep}.
However, most studies adopt a single UQ method without systematic evaluation, leaving the impact of specific design choices largely unexplored.

In this work, we present \emph{RewardUQ}, a unified framework for the design and evaluation of uncertainty-aware reward models.
This framework represents a first step towards principled UQ in preference modeling with reward functions, with the goal to build a foundation for more reliable and sample-efficient RLHF.
Our main contributions are as follows:
\begin{itemize}
    \item We introduce a unified framework which formalizes the UQ problem, standardizes existing methods in a consistent notation, and defines a common evaluation procedure.
    Our evaluations utilize a new ranking strategy that incorporates the accuracy and calibration of predictions under uncertainty.
    \item We conduct a systematic evaluation of existing uncertainty-aware reward models, analyzing how architectural choices and training parameters affect the quality of uncertainty estimates.
    \item We release an open-source Python package as an accessible and extensible platform for developing, evaluating, and deploying new UQ methods.
\end{itemize}

\section{Related work}\label{sec:related}

\subsection{Methods for reward model uncertainty quantification}


\paragraph{Ensembles}
\looseness=-1
The predominant approach for uncertainty quantification for reward models in RLHF are ensembles, with the uncertainty represented by the variance across ensemble members.
In its simplest form, an ensemble combines multiple reward models trained with different random seeds \citep{coste2023reward, eisenstein2024helping, liang2021reward, christiano2017deep} and, optionally, with bootstrapped datasets \citep{lou2025uncertaintyaware, gleave2022uncertainty}.
To reduce computational cost, others utilize a pretrained model and train only lightweight ensemble members such as linear heads \citep{banerjee2024reliable, yan2024rewardrobust}, multi-layer perceptron (MLP) heads \citep{melo2024deep, liu2024sampleefficient, dwaracherla2024efficient}, low-rank adaptation (LoRA) adapters \citep{zhai2024uncertaintypenalized, sun2024uncertainty, houliston2024uncertaintypenalized, yang2024bayesian, zhang2024improving}, or apply Monte Carlo (MC) dropout \citep{mehta2025sample, zhang2025dorm}.

\paragraph{Bayesian inference with Laplace approximation}
\looseness=-1
An alternative approach assumes a Gaussian prior on the parameters of a single reward model and derives the uncertainty from the predictive posterior based on the Laplace approximation \citep{cercola2025efficient}.
As the Hessian is often intractable over all model parameters, the Laplace approximation is typically applied to a subset of the parameters of a pretrained LLM, such as the linear head \citep{das2024active, cercola2025efficient} or a LoRA adapter \citep{yang2024bayesian}.
    
\paragraph{Mean-variance estimation}
Some studies utilize reward models which predict the mean and variance of a Gaussian reward distribution, capturing the aleatoric uncertainty under heteroscedastic noise \citep{yan2024rewardrobust, lou2025uncertaintyaware, siththaranjan2023distributional, sun2025probabilistic}.

\paragraph{Reward-margin-based preference uncertainty}
Others leverage the margin between pointwise rewards as a measure of uncertainty about the true preference, without modeling the uncertainty about the true reward \citep{muldrew2024active, lou2025uncertaintyaware}.

Our work focuses on the most common approaches and covers a selection of ensemble and Bayesian inference methods, identifying commonalities and differences, and evaluating them side by side.

\subsection{Applications for reward UQ}

\paragraph{Uncertainty-aware alignment}
Uncertainty estimates can make the alignment step in RLHF more resilient to reward overoptimization by encouraging the LLM to avoid uncertain rewards.
Common schemes involve penalizing \citep{lou2025uncertaintyaware, zhai2024uncertaintypenalized, houliston2024uncertaintypenalized, banerjee2024reliable, coste2023reward, sun2025probabilistic} or filtering \citep{sun2024uncertainty, lou2025uncertaintyaware} uncertain rewards.
Other approaches adopt pessimistic objectives to optimize for worst-case performance under uncertainty \citep{zhang2024mitigating, yan2024rewardrobust} or apply pessimistic best-of-$N$ sampling~\citep{liu2025uncertainty}.
    
\paragraph{Active learning for reward modeling}
To reduce the cost of collecting high-quality preference data, uncertainty estimates can guide the label acquisition towards more informative samples, improving the sample efficiency in the reward modeling step in RLHF \citep{melo2024deep, dwaracherla2024efficient, das2024active}.
Others utilize uncertainty to estimate the quality of and adaptively assign weights to preference samples \citep{zhang2025dorm}.
    
\paragraph{Active learning for alignment}
Similarly, uncertainty in the predicted rewards can improve the sample efficiency in the alignment step in RLHF, be it through uncertainty-based selection criteria of alignment samples \citep{mehta2025sample, muldrew2024active, christiano2017deep, cercola2025efficient} or exploration bonuses \citep{liu2024sampleefficient, liang2021reward}.

These directions highlight the promise of UQ methods for reward models.
Yet, most studies adopt a single method and focus on downstream applications. Even work that compare multiple methods, such as the ensemble architecture study of \citet{zhang2024improving}, limit their analysis to downstream performance rather than a systematic analysis of the uncertainty quantification itself.
In contrast, our work follows a complementary direction by focusing on the design and evaluation of different UQ methods, aiming to provide a clear comparison and offer insights on how to choose and use methods.
With most prior work initializing their reward models from generic pre-trained models, our results suggest that most works could have benefited from better design choices, especially by choosing model initializations that are tuned for reward modeling.
\section{Uncertainty quantification for reward models}\label{sec:uq}

We introduce a unified framework for designing and evaluating uncertainty-aware reward models, which integrates a range of existing approaches and extends them with novel contributions of our own.
We begin by formalizing the UQ problem for reward models in \cref{sec:uq-problem}, and then introduce our evaluation metrics in \cref{sec:uq-accuracy,sec:uq-calibration}.

\subsection{Problem statement}\label{sec:uq-problem}

We consider the reinforcement learning from human feedback (RLHF) problem, which aims to align a language model $\pi$ with human preferences, such that $\pi$ is more likely to generate a human-preferred completion $y\sim\pi(\cdot\mid x)$ for a given prompt $x$ \citep{ouyang2022training, stiennon2020learning}.
We assume preferences to be expressed as pairwise comparisons in terms of $(x,\chosen{y},\rejected{y})$ with $\chosen{y}$ being preferred over $\rejected{y}$, denoted $\chosen{y} \succ \rejected{y}$.
As standard in the literature, we assume the Bradley-Terry preference model \citep{bradley1952rank} that models the comparison between two candidate completions $y$ and $y'$ as a Bernoulli distribution with probability
\begin{equation}\label{eq:bt_model}
    p(y\succ y' \mid x,y,y') = \sigmoid*{r(x,y) - r(x,y')}
\end{equation}
where $\sigma(x) = \frac{1}{1+\exp(-x)}$ is the sigmoid function and $r$ is a reward function assigning a scalar score to any prompt-completion pair.
Given a dataset $\Dcal_{\text{train}} = \set{(x_i, \chosen{y}_i, \rejected{y}_i)}_{i=1}^n$, a reward model $\rmodel$ is trained by maximizing the likelihood of the observed preferences or equivalently by minimizing the binary cross-entropy loss
\begin{equation}\label{eq:loss-base}
    \Lcal_{\text{base}}(\theta;\Dcal_{\text{train}}) = \frac{1}{n} \sum_{(x,\chosen{y},\rejected{y}) \in \Dcal_{\text{train}}} -\log\sigmoid*{\rmodel(x,\chosen{y}) - \rmodel(x,\rejected{y})}
    .
\end{equation}
Once trained, $\rmodel$ can be used to align $\pi$ via RL algorithms such as PPO \citep{schulman2017proximal} or GRPO \citep{shao2024deepseekmath}, or at inference time with best-of-$N$ sampling \citep{stiennon2020learning, yang2024bayesian}.
However, the standard RLHF framework relies on reward models which only make pointwise predictions, thereby neglecting the epistemic uncertainty arising from training on a finite dataset sampled from the large domain of natural language.

An uncertainty-aware reward model additionally predicts upper and lower confidence bounds $\ub{\rmodel}(x,y)$ and $\lb{\rmodel}(x,y)$, quantifying its epistemic uncertainty about the true underlying reward in terms of a confidence interval $\interval{\rmodel}(x,y) = \brack*{\lb{\rmodel}(x,y), \ub{\rmodel}(x,y)}$.
We introduce the most common methods in detail in \cref{sec:models}.
Under the Bradley-Terry model assumption, the corresponding upper and lower bounds on the preference probability are given by
\begin{equation}\label{eq:preferences-bounds}
    \begin{aligned}
        \ub{\pmodel}(y\succ y' \mid x,y,y') &= \sigmoid*{\ub{\rmodel}(x,y) - \lb{\rmodel}(x,y')}
        \\ \lb{\pmodel}(y\succ y' \mid x,y,y') &= \sigmoid*{\lb{\rmodel}(x,y) - \ub{\rmodel}(x,y')}
        ,
    \end{aligned}
\end{equation}
which are based on the largest and smallest plausible reward margin between both candidate completions, respectively~\citep{mehta2025sample}.

The goal of an uncertainty-aware reward model is to predict preference probabilities and confidence bounds, which not only accurately reflect the true binary preferences but are also statistically well-calibrated with respect to the true preference probabilities.
We introduce our precise notion of accuracy and calibration, two complementary evaluation dimensions, along with their corresponding metrics in \cref{sec:uq-accuracy,sec:uq-calibration}.
We elaborate on the epistemic and aleatoric uncertainty decomposition for preference classification in \cref{sec:theoretical_details-uncertainty_decomposition}, and discuss the differences between standard and preference classification by focusing on the symmetry of \cref{eq:bt_model} in \cref{sec:theoretical_details-preference_classification}.

\subsection{Accuracy metrics}\label{sec:uq-accuracy}

While accuracy is a standard performance measure for pointwise predictions, we further extend the notion of accuracy to confidence bounds.

\paragraph{Accuracy of predictions}
Given an evaluation dataset $\Dcal_{\text{eval}}$ and a reward model $\rmodel$, the predicted rewards are correct if they assign higher rewards to preferred completions.%
\footnote{This is equivalent to $\pmodel(\chosen{y}\succ\rejected{y}\mid x,\chosen{y},\rejected{y}) > 0.5$ under the Bradley-Terry model in \cref{eq:bt_model}.}
Let
\begin{equation*}
    \begin{aligned}
        \mathbf{T}\mathrm{(rue)} &= \set{(x,\chosen{y},\rejected{y}) \in \Dcal_{\text{eval}} \mid \rmodel(x,\chosen{y}) > \rmodel(x,\rejected{y})}
        \\ \mathbf{F}\mathrm{(alse)} &= \set{(x,\chosen{y},\rejected{y}) \in \Dcal_{\text{eval}} \mid \rmodel(x,\chosen{y}) \leq \rmodel(x,\rejected{y})}
    \end{aligned}
\end{equation*}
be the set of true (\ie{} correct) and false (\ie{} incorrect) preference predictions.
The accuracy, commonly known as \emph{win rate} in the context of RLHF, is defined as
\begin{equation}\label{eq:metrics-accuracy-predictions}
    \begin{alignedat}{2}
        \mathrm{win\ rate} &= \frac{\abs{\mathbf{T}}}{n}.  && \quad\uparrow\footnotemark
    \end{alignedat}
\end{equation}
\footnotetext{The arrow indicates the direction of improvement in the metric value.}\vspace{-\baselineskip}

\paragraph{Accuracy of bounds}
While the win rate only captures the accuracy of pointwise predictions, we extend this notion of accuracy to confidence intervals.
To quantify the accuracy of the predicted reward confidence intervals $\interval{\rmodel}(x,y)$, we further categorize the true and false predictions into
\begin{equation*}
    \begin{aligned}
        \mathbf{C}\mathrm{(onfident)} &= \set{(x,\chosen{y},\rejected{y}) \in \Dcal_{\text{eval}} \mid \interval{\rmodel}(x,\chosen{y}) \cap \interval{\rmodel}(x,\rejected{y}) = \emptyset}
        \\ \mathbf{U}\mathrm{(nconfident)} &= \set{(x,\chosen{y},\rejected{y}) \in \Dcal_{\text{eval}} \mid \interval{\rmodel}(x,\chosen{y}) \cap \interval{\rmodel}(x,\rejected{y}) \neq \emptyset}
        .
    \end{aligned}
\end{equation*}
Intuitively, a prediction is confident when the predicted reward confidence intervals of the preferred and non-preferred completion do not overlap, indicating no ambiguity in the predicted preference even under uncertainty.
By combining the correctness of the pointwise predictions with the confidence of the predicted bounds, we define the following metrics
\begin{equation}\label{eq:metrics-accuracy-bounds}
    \begin{alignedat}{5}
        &
        && \eqnote{confident}
        &&
        && \eqnote{unconfident}
        \\ \eqnote{true} &
        \qquad
        & \mathrm{CT\ rate} &= \frac{\abs{\mathbf{C} \cap \mathbf{T}}}{n}
        && \quad\uparrow
        \qquad
        & \mathrm{UT\ rate} &= \frac{\abs{\mathbf{U} \cap \mathbf{T}}}{n}
        && \quad\searrow
        \\ \eqnote{false} &
        \qquad
        & \mathrm{CF\ rate} &= \frac{\abs{\mathbf{C} \cap \mathbf{F}}}{n}
        && \quad\downarrow
        \qquad
        & \mathrm{UF\ rate} &= \frac{\abs{\mathbf{U} \cap \mathbf{F}}}{n}
        .
        && \quad\searrow
    \end{alignedat}
\end{equation}
\looseness=-1
We refer to \cref{sec:theoretical_details-accuracy} for a generalization of these metrics to the standard binary classification setting.

\paragraph{Ranking score}
In order to compare models, we propose a ranking score that combines the accuracy metrics above into a single score.
Motivated by the UQ reward model applications, this score encourages a high confident true rate to efficiently guide active learning algorithms and identify reliable training data samples.
Simultaneously, it penalizes the confident false rate that could provide misleading signals.
The ranking score is defined as
\begin{equation}\label{eq:metrics-accuracy-ranking}
    \begin{aligned}
        \mathrm{RS}_\alpha
        &= \frac{\mathrm{CT\ rate}}{\mathrm{win\ rate} + \alpha \cdot (1 - \mathrm{win\ rate})} - \frac{\mathrm{CF\ rate}}{(1 - \mathrm{win\ rate}) + \alpha \cdot \mathrm{win\ rate}}
        \\ &= \frac{\abs{\mathbf{C} \cap \mathbf{T}}}{\abs{\mathbf{T}} + \alpha \cdot \abs{\mathbf{F}}} - \frac{\abs{\mathbf{C} \cap \mathbf{F}}}{\abs{\mathbf{F}} + \alpha \cdot \abs{\mathbf{T}}}
        \in [-1, 1]
    \end{aligned}
    \quad\uparrow
\end{equation}
using a trade-off parameter $\alpha\in[0,1]$, which balances the focus on the confidence and the focus on the accuracy.
For $\alpha=0$, $\mathrm{RS}_0$ considers the relative rate of confidence among true and false predictions by normalizing the $\mathrm{CT}$ and $\mathrm{CF}$ rates and represents performance in the range $[-1, 1]$.
In other words, high $\mathrm{RS}_0$ scores are achievable without achieving a high win rate.
For $\alpha=1.0$, $\mathrm{RS}_1$ looks at the absolute difference between the confident true and false predictions and represents performance in the range $[\mathrm{win\ rate} - 1, \mathrm{win\ rate}]$.
This approach assigns equal weight to $\mathrm{CT}$ and $\mathrm{CF}$ and favors a high win rate over confidence.
For our evaluations in \cref{sec:experiments}, we choose $\mathrm{RS}_{0.2}$ as a balance between accuracy and confidence.
A more detailed explanation of the rationale and analysis of the inherent trade-offs are provided in \cref{sec:theoretical_details-ranking}.

\subsection{Calibration metrics}\label{sec:uq-calibration}

Calibration refers to the gap between a predicted probability and the true (or empirical) probability.
We give a formal description of calibration in \cref{sec:theoretical_details-calibration} while providing an overview here.

\paragraph{Calibration of predictions}
The \emph{expected calibration error (ECE)} is commonly used to measure the calibration of predicted preference probabilities \citep{zhai2024uncertaintypenalized, gleave2022uncertainty}.
It is approximated based on grouping the predicted probabilities into $M$ bins $\set{B_m}_{m=1}^M$ and computing
\begin{equation}\label{eq:metrics-calibration-predictions}
     \begin{alignedat}{3}
        \mathrm{ECE} &\approx \sum_{m=1}^{M} \frac{\abs{B_m}}{n} \abs*{\ptrue(B_m) - \pmodel(B_m)} && \quad\downarrow
    \end{alignedat}
\end{equation}
with empirical probability $\ptrue(B_m)$ and average predicted probability $\pmodel(B_m)$ in each bin $B_m$ \citep{guo2017calibration, pavlovic2025understanding}.

\paragraph{Calibration of bounds}
We extend the notion of calibration to the predicted preference probability bounds constructed in \cref{eq:preferences-bounds} by introducing the \emph{expected lower calibration error (ELCE)} and \emph{expected upper calibration error (EUCE)}.
Analog to ECE, we group the lower and upper bounds separately into $M$ bins $\set{B_m}_{m=1}^M$ and compute
\begin{equation*}
    \begin{alignedat}{3}
        \mathrm{ELCE} &\approx \sum_{m=1}^{M} \frac{\abs{B_m}}{n} \max\paren*{\lb{\pmodel}(B_m) - \ptrue(B_m), 0} && \quad\downarrow
        \\ \mathrm{EUCE} &\approx \sum_{m=1}^{M} \frac{\abs{B_m}}{n} \max\paren*{\ptrue(B_m) - \ub{\pmodel}(B_m), 0} && \quad\downarrow
    \end{alignedat}
\end{equation*}
with $\lb{\pmodel}(B_m)$ and $\ub{\pmodel}(B_m)$ denoting the average predicted lower and upper bounds in the corresponding bin $B_m$.
ELCE penalizes lower bounds that overestimate the true preference probability, and EUCE penalizes upper bounds that underestimate the true preference probability.

Note that preference probabilities are antisymmetric in their completions argument. Accordingly, all calibration metrics are computed on a symmetrized preference evaluation set that includes flipped comparisons with opposite labels, ensuring that both directions of each preference pair contribute to the binning-based approximation of the calibration errors, as further described in \cref{sec:theoretical_details-preference_classification}.
Therefore, the lower bound on the probability of $y\succ y'$ corresponds at the same time to an upper bound for the probability of $y'\succ y$.
Hence, ELCE and EUCE are identical in the context of preference probabilities, and we subsequently only report the \emph{expected bound calibration error (EBCE)}
\begin{equation}\label{eq:metrics-calibration-bounds}
    \begin{alignedat}{3}
        \mathrm{EBCE} &= \mathrm{ELCE} = \mathrm{EUCE}
        .
        && \quad\downarrow
    \end{alignedat}
\end{equation}

\section{Uncertainty-aware reward models}\label{sec:models}

\begin{figure}
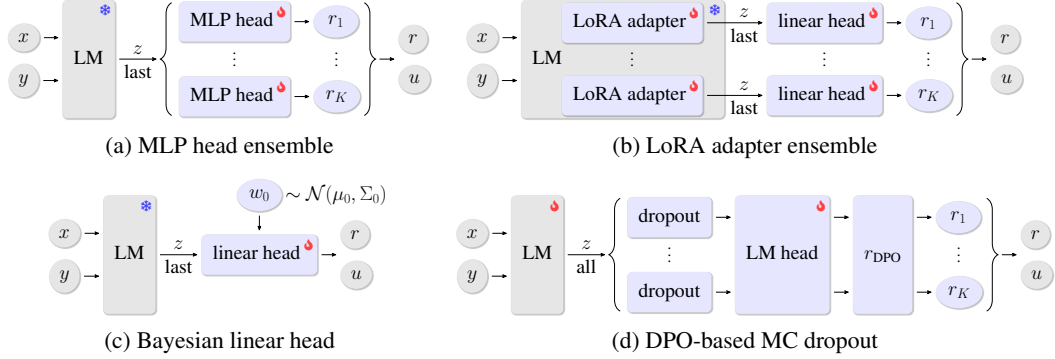

    \centering
    \begin{subfigure}{0.45\linewidth}
        \centering
        \scalebox{0.3}{\begin{tikzpicture}
\input{tikz/style}

\node[var, above=0.25cm] (x) {\(x\)};
\node[var, below=0.25cm] (y) {\(y\)};
\node[module, right=of $(x.east)!0.5!(y.east)$, minimum height=5.25cm] (model) {LM};
\drawsymbol{model}{\freezesymbol}

\draw[arrow] (x) -- ($(model.north west)!(x)!(model.south west)$);
\draw[arrow] (y) -- ($(model.north west)!(y)!(model.south west)$);

\coordinate[right=2cm of model] (start);
\draw[arrow] (model) -- (start) 
    node[label, midway, above=0.1cm] {\(z\)}
    node[label, midway, below=0.1cm] {last};

\node[module, focus, above right=0.75cm and 0.8cm of start] (head1) {MLP head\thickspace};
\node[module, focus, below right=0.75cm and 0.8cm of start] (head2) {MLP head\thickspace};
\drawsymbol{head1}{\trainsymbol}
\drawsymbol{head2}{\trainsymbol}
\node[var, focus, right=of head1, minimum width=2cm] (r1) {\(r_1\)};
\node[var, focus, right=of head2, minimum width=2cm] (r2) {\(r_K\)};

\node (dots) at ($(head1)!0.5!(head2)$) [yshift=0.2cm] {\scalebox{2}{\(\vdots\)}};
\node (dots2) at ($(r1)!0.5!(r2)$) [yshift=0.2cm] {\scalebox{2}{\(\vdots\)}};

\draw[arrow] (head1) -- (r1);
\draw[arrow] (head2) -- (r2);

\coordinate[right=0.7cm of $(r1.east)!0.5!(r2.east)$] (end);

\drawbraceopen{start}{2.5cm}
\drawbraceclose{end}{2.5cm}

\node[var, right=of end, anchor=ext_corner south west, yshift=0.25cm] (r) {\(r\)};
\node[var, right=of end, anchor=ext_corner north west, yshift=-0.25cm] (sigma) {\(u\)};

\draw[arrow] (end) -- ($(r.west)!0.5!(sigma.west)$);

\end{tikzpicture}}
        \caption{MLP head ensemble}
        \label{fig:models-mlp_ensemble}
    \end{subfigure}%
    \begin{subfigure}{0.55\linewidth}
        \centering
        \scalebox{0.3}{\begin{tikzpicture}
\input{tikz/style}

\node[var, above=0.25cm] (x) {\(x\)};
\node[var, below=0.25cm] (y) {\(y\)};

\node[module, right=of $(x.east)!0.5!(y.east)$, minimum height=5.25cm, text width=8cm, align=left] (model) {LM};
\drawsymbol{model}{\freezesymbol}

\draw[arrow] (x) -- ($(model.north west)!(x)!(model.south west)$);
\draw[arrow] (y) -- ($(model.north west)!(y)!(model.south west)$);


\node[module, focus, above right=0.7cm and 1.8cm of model.west] (adapter1) {LoRA adapter\thickspace};
\node[module, focus, below right=0.7cm and 1.8cm of model.west] (adapter2) {LoRA adapter\thickspace};
\drawsymbol{adapter1}{\trainsymbol}
\drawsymbol{adapter2}{\trainsymbol}

\node[module, focus, right=2.8cm of adapter1] (head1) {linear head\thickspace};
\node[module, focus, right=2.8cm of adapter2] (head2) {linear head\thickspace};
\drawsymbol{head1}{\trainsymbol}
\drawsymbol{head2}{\trainsymbol}

\draw[arrow] (adapter1) -- (head1)
    node[label, pos=0.65, above=0.1cm] {\(z\)}
    node[label, pos=0.65, below=0.1cm] {last};
    
\draw[arrow] (adapter2) -- (head2)
    node[label, pos=0.65, above=0.1cm] {\(z\)}
    node[label, pos=0.65, below=0.1cm] {last};

\node[var, focus, right=of head1, minimum width=2cm] (r1) {\(r_1\)};
\node[var, focus, right=of head2, minimum width=2cm] (r2) {\(r_K\)};

\draw[arrow] (head1) -- (r1);
\draw[arrow] (head2) -- (r2);

\coordinate[right=0.7cm of $(r1.east)!0.5!(r2.east)$] (end);
\drawbraceclose{end}{2.5cm}

\node (dots) at ($(adapter1)!0.5!(adapter2)$) [yshift=0.2cm] {\scalebox{2}{\(\vdots\)}};
\node (dots) at ($(head1)!0.5!(head2)$) [yshift=0.2cm] {\scalebox{2}{\(\vdots\)}};
\node (dots2) at ($(r1)!0.5!(r2)$) [yshift=0.2cm] {\scalebox{2}{\(\vdots\)}};

\node[var, right=of end, anchor=ext_corner south west, yshift=0.25cm] (r) {\(r\)};
\node[var, right=of end, anchor=ext_corner north west, yshift=-0.25cm] (sigma) {\(u\)};

\draw[arrow] (end) -- ($(r.west)!0.5!(sigma.west)$);

\end{tikzpicture}}
        \caption{LoRA adapter ensemble}
        \label{fig:models-lora_ensemble}
    \end{subfigure}%
    \par\medskip
    \begin{subfigure}{0.45\linewidth}
        \centering
        \scalebox{0.3}{\begin{tikzpicture}
\input{tikz/style}

\node[var, above=0.25cm] (x) {\(x\)};
\node[var, below=0.25cm] (y) {\(y\)};
\node[module, right=of $(x.east)!0.5!(y.east)$, minimum height=5.25cm] (model) {LM};
\drawsymbol{model}{\freezesymbol}

\draw[arrow] (x) -- ($(model.north west)!(x)!(model.south west)$);
\draw[arrow] (y) -- ($(model.north west)!(y)!(model.south west)$);

\coordinate[right=2cm of model] (start);
\draw[arrow] (model) -- (start) 
    node[label, midway, above=0.1cm] {\(z\)}
    node[label, midway, below=0.1cm] {last};

\node[module, focus, right=0cm of start] (head) {linear head\thickspace};
\drawsymbol{head}{\trainsymbol}
\node[var, focus, above=of head] (initial) {\(w_0\)};
\node[right=0cm of initial, font=\normalsize] (prior) {\(\sim\mathcal{N}(\mu_0,\Sigma_0)\)};

\draw[arrow] (initial) -- (head);

\coordinate[right=0cm of head] (end);

\node[var, right=of end, anchor=ext_corner south west, yshift=0.25cm] (r) {\(r\)};
\node[var, right=of end, anchor=ext_corner north west, yshift=-0.25cm] (sigma) {\(u\)};

\draw[arrow] (end) -- ($(r.west)!0.5!(sigma.west)$);

\end{tikzpicture}}
        \caption{Bayesian linear head}
        \label{fig:models-bayesian_linear}
    \end{subfigure}%
    \begin{subfigure}{0.55\linewidth}
        \centering
        \scalebox{0.3}{\begin{tikzpicture}
\input{tikz/style}

\node[var, above=0.25cm] (x) {\(x\)};
\node[var, below=0.25cm] (y) {\(y\)};
\node[module, right=of $(x.east)!0.5!(y.east)$, minimum height=5.25cm] (model) {LM};
\drawsymbol{model}{\trainsymbol}

\draw[arrow] (x) -- ($(model.north west)!(x)!(model.south west)$);
\draw[arrow] (y) -- ($(model.north west)!(y)!(model.south west)$);

\coordinate[right=2cm of model] (start);
\draw[arrow] (model) -- (start) 
    node[label, midway, above=0.1cm] {\(z\)}
    node[label, midway, below=0.1cm] {all};

\node[module, focus, above right=0.75cm and 0.8cm of start] (dropout1) {dropout};
\node[module, focus, below right=0.75cm and 0.8cm of start] (dropout2) {dropout};
\node[module, focus, right=of $(dropout1.east)!0.5!(dropout2.east)$, minimum height=5.25cm] (head) {\(\text{LM head}\)};
\drawsymbol{head}{\trainsymbol}
\node[module, focus, right=of head, minimum height=5.25cm] (dpo) {\(r_{\text{DPO}}\)};
\node[var, focus, right=of $(dpo.north east)!(dropout1)!(dpo.south east)$, minimum width=2cm] (r1) {\(r_1\)};
\node[var, focus, right=of $(dpo.north east)!(dropout2)!(dpo.south east)$, minimum width=2cm] (r2) {\(r_K\)};

\node (dots) at ($(dropout1)!0.5!(dropout2)$) [yshift=0.2cm] {\scalebox{2}{\(\vdots\)}};
\node (dots2) at ($(r1)!0.5!(r2)$) [yshift=0.2cm] {\scalebox{2}{\(\vdots\)}};

\draw[arrow] (dropout1) -- ($(head.north west)!(dropout1)!(head.south west)$);
\draw[arrow] (dropout2) -- ($(head.north west)!(dropout2)!(head.south west)$);
\draw[arrow] ($(head.north east)!(dropout1)!(head.south east)$) -- ($(dpo.north west)!(r1)!(dpo.south west)$);
\draw[arrow] ($(head.north east)!(dropout2)!(head.south east)$) -- ($(dpo.north west)!(r2)!(dpo.south west)$);
\draw[arrow] ($(dpo.north east)!(r1)!(dpo.south east)$) -- (r1);
\draw[arrow] ($(dpo.north east)!(r2)!(dpo.south east)$) -- (r2);

\coordinate[right=0.7cm of $(r1.east)!0.5!(r2.east)$] (end);

\drawbraceopen{start}{2.5cm}
\drawbraceclose{end}{2.5cm}

\node[var, right=of end, anchor=ext_corner south west, yshift=0.25cm] (r) {\(r\)};
\node[var, right=of end, anchor=ext_corner north west, yshift=-0.25cm] (sigma) {\(u\)};

\draw[arrow] (end) -- ($(r.west)!0.5!(sigma.west)$);

\end{tikzpicture}}
        \caption{DPO-based MC dropout}
        \label{fig:models-dpo_mcd}
    \end{subfigure}
    \caption{
        Uncertainty-aware reward model architectures compared in this work.
        For a given prompt $x$ and completion $y$, each model extracts an embedding $z$ from a pretrained language model (LM) and predicts a reward $r$ and uncertainty estimate $u$.
        Blue components indicate the parts responsible for estimating the uncertainty, while \trainsymbol{} and \freezesymbol{} denote trainable and frozen components, respectively.
    }
    \label{fig:models}
\end{figure}

In this work, we focus on the most common uncertainty-aware reward model architectures from existing work.
While these models differ in how they represent epistemic uncertainty, they share several core principles as illustrated on \cref{fig:models}.
Following prior work~\citep{li2022nearoptimal, ji2024reinforcement, mehta2025sample}, for any prompt-completion pair, $(x,y)$, we separate pointwise prediction from uncertainty quantification, and assume that each model predicts a reward $\rmodel(x,y)$ and an uncertainty estimate $\umodel(x,y)$, which are used to construct the symmetric confidence bounds
\begin{equation}\label{eq:reward-bounds}
    \begin{aligned}
        \ub{\rmodel}(x,y) &= \rmodel(x,y) + \beta\cdot\umodel(x,y)
        \\ \lb{\rmodel}(x,y) &= \rmodel(x,y) - \beta\cdot\umodel(x,y)
    \end{aligned}
\end{equation}
with scaling factor $\beta > 0$.
The reward models are trained using the standard binary cross-entropy loss defined in \cref{eq:loss-base}, with modifications depending on the specific architecture as described below.

\subsection{MLP head ensemble (ENS-MLP)}\label{sec:models-mlp_ensemble}
\looseness=-1
A common approach to estimate epistemic uncertainty is to train an ensemble of $K$ independent Multi-Layer Perceptron (MLP) heads using the embedding $z$ provided by a pretrained LLM \citep{melo2024deep, liu2024sampleefficient, dwaracherla2024efficient}, as illustrated in \cref{fig:models-mlp_ensemble}.
Each MLP head is parametrized by $\theta^{(k)} \in \Real^d$ and predicts a pointwise reward $\rmodel[^{(k)}](x,y)$.
The pointwise reward and uncertainty estimates in \cref{eq:reward-bounds} are computed as the mean and standard deviation over the individual rewards
\begin{equation}\label{eq:models-ens_mlp-estimates}
    \rmodel(x,y) = \frac{1}{K} \sum_{k=1}^K \rmodel[^{(k)}](x,y)
    \quad\text{and}\quad
    \umodel(x,y) = \sqrt{\frac{1}{K-1} \sum_{k=1}^K \paren*{\rmodel[^{(k)}](x,y) - \rmodel(x,y)}^2}.
\end{equation}
The model is trained by minimizing the loss
{\small
\begin{equation}\label{eq:models-ens_mlp-loss}
    \begin{split}
        \Lcal(\theta;\Dcal_{\text{train}}) &= \frac{1}{K} \sum_{k=1}^K \paren[\bigg]{ \Lcal_{\text{base}}(\theta^{(k)};\Dcal_{\text{train}}) + \frac{\lambda}{d} \norm*{\theta^{(k)} - \theta^{(k)}_{\text{init}}}_2^2 \\
        &+ \frac{\gamma}{n} \sum_{(x,\chosen{y},\rejected{y}) \in \Dcal_{\text{train}}} (\rmodel[^{(k)}](x,\chosen{y}) + \rmodel[^{(k)}](x, \rejected{y}))^2 }
        ,
    \end{split}
\end{equation}
}%
which consists of the standard cross-entropy loss from \cref{eq:loss-base} and two regularization terms applied on each head.
The first regularization term controlled by $\lambda$ encourages the parameters of each head $\theta^{(k)}$ to stay close to their random initialization $\theta^{(k)}_{\text{init}}$, which preserves diversity across the heads in the ensemble.
The second regularization term controlled by $\gamma$ centers predicted rewards around zero \citep{eisenstein2024helping}.
This is a crucial practical step often overseen in practice, as the cross-entropy loss is invariant to additive constants in the reward function, which could otherwise lead to poorly calibrated uncertainty estimates due to unintended large standard deviations.

\subsection{LoRA adapter ensemble (ENS-LoRA)}\label{sec:models-lora_ensemble}


\looseness=-1
ENS-LoRA~\citep{muhlematter2025loraensemble} extends the framework of the MLP head ensemble model defined in \cref{sec:models-mlp_ensemble} by training all layers of the model instead of additional MLP heads.
To overcome the computational constraints of training $K$ models, Low-Rank Adaptation (LoRA) method is used to reduce the number of trainable parameters \citep{wang2023lora}.
We denote each LoRA adapter by the parameter vector $\theta^{(k)}$ and initialize a linear head for each adapter to obtain a pointwise reward $\rmodel[^{(k)}](x,y)$ from embeddings $z$. The adapters are trained by minimizing the loss defined in \cref{eq:models-ens_mlp-loss} and the reward $\rmodel(x,y)$ and uncertainty estimates $\umodel(x,y)$ are computed following \cref{eq:models-ens_mlp-estimates}.

\subsection{DPO-based MC dropout (MCD-DPO)}\label{sec:models-dpo_mcd}

Instead of training several heads of LoRA adapters, one can also leverage Monte-Carlo (MC) dropouts before the final layer of a fine-tuned model and estimate rewards implicitly.
Formally, let $\pi_\theta$ be a fine-tuned LLM initialized from a reference policy $\pi_{\text{ref}}$ and trained to minimize the KL-regularized loss~\citep{christiano2017deep, ouyang2022training, stiennon2020learning}.
This policy defines an implicit reward model as
\begin{equation}\label{eq:dpo_mcd-implicit_reward_model}
    \rmodel(x,y) = \lambda \log{\frac{\pi_\theta(y\mid x)}{\pi_{\text{ref}}(y\mid x)}} + \lambda \log{Z(x)}
    ,
\end{equation}
where $\lambda$ controls the KL-regularization term and $Z(x)$ is the partition function~\citep{rafailov2023direct}.
MCD-DPO~\citep{mehta2023sample} quantifies the uncertainty of this implicit reward function by introducing a dropout layer right before the language modeling head to enable MC dropout~\citep{gal2016dropout}, as shown in \cref{fig:models-dpo_mcd}.

During inference, $K$ dropout masks $m^{(k)}$ are sampled and applied to the embedding $z$ of a pretrained LLM, providing an ensemble of completion probabilities $\pi_\theta(y\mid x; m^{(k)})$.
We obtain the implicit rewards $\rmodel(x,y; m^{(k)})$ using \cref{eq:dpo_mcd-implicit_reward_model}, denoted by the $r_{\text{DPO}}$ layer in \cref{fig:models-dpo_mcd}, and utilize the mean and standard deviation over these individual rewards
\begin{equation*}
    \rmodel(x,y) = \frac{1}{K} \sum_{k=1}^K \rmodel(x,y; m^{(k)})
    \quad\text{and}\quad
    \umodel(x,y) = \sqrt{\frac{1}{K-1} \sum_{k=1}^K \paren*{\rmodel(x,y; m^{(k)}) - \rmodel(x,y)}^2}
\end{equation*}
as our reward and uncertainty estimate for \cref{eq:reward-bounds}.
In our implementation $\pi_\theta$ is trained with the DPO loss, derived by substituting \cref{eq:dpo_mcd-implicit_reward_model} into the standard cross-entropy loss in \cref{eq:loss-base},
\begin{equation}\label{eq:models-dpo_mcd-loss}
    \Lcal(\theta;\Dcal_{\text{train}}) = \frac{1}{n} \sum_{\substack{(x,\chosen{y},\rejected{y}) \in \Dcal_{\text{train}} \\ m\sim\P_\text{dropout}}} -\log\sigmoid*{\lambda \log{\frac{\pi_\theta(\chosen{y}\mid x; m)}{\pi_{\text{ref}}(\chosen{y}\mid x)}} - \lambda \log{\frac{\pi_\theta(\rejected{y}\mid x; m)}{\pi_{\text{ref}}(\rejected{y}\mid x)}}}
    ,
\end{equation}
with a randomly sampled dropout mask $m$ per sample.

\subsection{Bayesian linear head (BAY-LIN)}\label{sec:models-bayesian_linear}

Another common approach in the literature is to consider reward estimation as a Bayesian linear regression problem~\citep{das2024active, cercola2025efficient}.
This method also computes the embedding $z$ for each prompt-completion pair $(x,y)$ but applies a single linear reward head
\begin{equation*}
    \rmodel(x,y) = \theta^\top z
\end{equation*}
with a Gaussian prior on the trainable parameters $\theta \sim \Normal*{0, \lambda^{-1} I}$.
The posterior on $\theta$ is then approximated using a Laplace approximation, resulting in the following Gaussian distribution
\begin{equation*}
     \theta\mid\Dcal_{\text{train}} \stackrel{\text{approx}}{\sim} \Normal*{\theta_{\text{MAP}},H^{-1} \big\vert_{\theta=\theta_{\text{MAP}}}}
\end{equation*}
with mean centered at the posterior mode $\theta_{\text{MAP}} = \argmin_{\theta} -\log{p(\theta|\Dcal_{\text{train}})}$ and the inverse covariance given by the Hessian of the negative log-posterior $H = \nabla^2_\theta -\log{p(\theta|\Dcal_{\text{train}})}$ evaluated at $\theta_{\text{MAP}}$.
Intuitively, the Gaussian distribution is centered at and fitted to the local curvature around the posterior mode.
The posterior mode is obtained by equivalently minimizing
\begin{equation}\label{eq:models-bay_lin-loss}
    \theta_{\text{MAP}} = \argmin_{\theta} \Lcal_{\text{base}}(\theta;\Dcal_{\text{train}}) + \frac{\lambda}{2} \norm{\theta}_2^2
\end{equation}
corresponding to the cross-entropy loss in \cref{eq:loss-base} with $\ell_2$-regularization.
The Hessian
\begin{equation}\label{eq:hessian-weighted}
    H = \sum_{(x,\chosen{y},\rejected{y}) \in \Dcal_{\text{train}}} 
    w(x,\chosen{y},\rejected{y}) \cdot 
    (\chosen{z} - \rejected{z})(\chosen{z} - \rejected{z})^\top + \lambda I,
\end{equation}
with weights $w(x,\chosen{y},\rejected{y}) = \sigmoid'\paren*{\rmodel(x,\chosen{y}) - \rmodel(x,\rejected{y})}$ corresponds to the empirical covariance of the feature differences with larger weights for ambiguous predictions, \ie{} $\rmodel(x,\chosen{y}) \approx \rmodel(x,\rejected{y})$.
However, these weights depend on the current parameter estimate $\theta$.
This dependence requires the entire sum in the Hessian to be recomputed in the active learning setting, where $\theta$ is updated iteratively.
To avoid these high computational costs, \citet{das2024active} omit these weights, allowing the Hessian $H$ to be updated incrementally.
We follow this unweighted approach to keep our evaluation practical.
The final reward and uncertainty estimate in \cref{eq:reward-bounds} are given by the predictive posterior mean and standard deviation
\begin{equation*}
    \rmodel(x,y) = \theta^\top z
    \quad\text{and}\quad
    \umodel(x,y) = \sqrt{z^\top H^{-1} z}
    .
\end{equation*}

\section{Experiments}\label{sec:experiments}

\begin{figure}
    \centering
    \includegraphics[width=\linewidth]{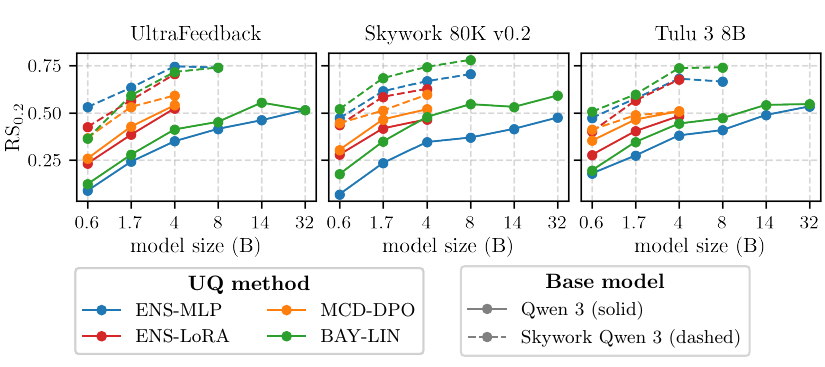}
    \caption{
        Ranking scores on RewardBench across different UQ methods, training datasets, pretrained and finetuned models, and model sizes.
        The ranking score is defined in \cref{eq:metrics-accuracy-ranking}.
    }
    \label{fig:results-ranking}
\end{figure}

For a systematic comparison, we train and evaluate the model architectures outlined in \cref{sec:models} across different datasets and base models with a unified evaluation procedure as described in \cref{sec:experiments-setup}.
Our main results are detailed in \cref{sec:experiments-results}.

\subsection{Experimental setup}\label{sec:experiments-setup}

\paragraph{Evaluation procedure}
For each uncertainty quantification method, we first perform a parameter optimization over common parameters (\eg{} learning rate, base model family, and model size) and model-specific ones (\eg{} regularization parameters, dropout rate).
This search is done on the training and validation split of the UltraFeedback preference dataset\footnote{\url{https://huggingface.co/datasets/trl-lib/ultrafeedback_binarized}} \citep{cui2024ultrafeedback} consisting of around 62K and 1K samples, respectively.
We select the best parameters by first applying an upper threshold on $\mathrm{ECE}$ and $\mathrm{EBCE}$ as introduced in \cref{eq:metrics-calibration-predictions,eq:metrics-calibration-bounds} with $0.05$ and $0.01$, respectively, to ensure reasonable calibration, and then ranking according to $\mathrm{RS}_{0.2}$\footnote{We describe our choice of $\alpha=0.2$ in \cref{sec:theoretical_details-ranking}.} from \cref{eq:metrics-accuracy-ranking}.
We report the final performance on the popular RewardBench dataset\footnote{We use the \texttt{filtered} split from \url{https://huggingface.co/datasets/allenai/reward-bench}} \citep{lambert2024rewardbench}.
Finally, we train each model on two additional datasets to evaluate our results robustness to the dataset's source and size: the Skywork preference dataset\footnote{\url{https://huggingface.co/datasets/Skywork/Skywork-Reward-Preference-80K-v0.2}} \citep{liu2024skyworkreward} with around 77K samples and the preference dataset for the Tulu 3 8B model\footnote{\url{https://huggingface.co/datasets/allenai/llama-3.1-tulu-3-8b-preference-mixture}} \citep{lambert2025tulu} with around 273K samples.

\paragraph{Models}
\looseness=-1
We initialize our models from either the Qwen 3 family\footnote{\url{https://huggingface.co/collections/Qwen/qwen3-67dd247413f0e2e4f653967f}} \citep{qwenteam2025qwen3} with sizes from 0.6B to 32B, and the Skywork-Reward-V2 Qwen 3 series\footnote{\url{https://huggingface.co/collections/Skywork/skywork-reward-v2-685cc86ce5d9c9e4be500c84}} \citep{liu2025skyworkrewardv2}, which are further finetuned for the reward modeling task on a large-scale preference dataset of around 26M preference pairs and range from 0.6B to 8B. This provides us with a broad coverage over model sizes and pre-training purposes.

Due to the larger computational requirements of ENS-LoRA and MCD-DPO, we consider models only up to 4B. Further experimental details, including hyperparameters, are provided in \cref{sec:experimental_details}.

\subsection{Results}\label{sec:experiments-results}

\paragraph{Insights into accuracy}
As illustrated in \cref{fig:results-ranking}, no single uncertainty quantification algorithm consistently dominates according to the ranking score $RS_{0.2}$; rather, performance is highly contingent on model size, dataset, and pre-training.
A critical determinant of performance is the base model initialization. Methods that rely on a fixed LLM backbone to provide embeddings, such as BAY-LIN and ENS-MLP, benefit significantly from initialization with a task-aligned reward model (e.g., the Skywork family). Conversely, when initialized from a generic base like Qwen 3, these methods underperform compared to ENS-LoRA and MCD-DPO, which fine-tune the full model parameters and are thus less sensitive to the quality of the initial embeddings.
Additionally, we observe diminishing returns in ranking scores as model size increases, a phenomenon we attribute to the higher overconfidence of larger models, which is penalized by our metric.
While BAY-LIN achieves the highest performance across most settings, it lags behind ENS-MLP on the UltraFeedback dataset, preventing a definitive recommendation. However, given that prior works typically utilize generic initializations, our findings strongly suggest that adopting task-aligned base models offers a potential for performance improvement.


\begin{figure}
    \begin{subfigure}{0.5\linewidth}
        \centering
        \includegraphics[width=\linewidth]{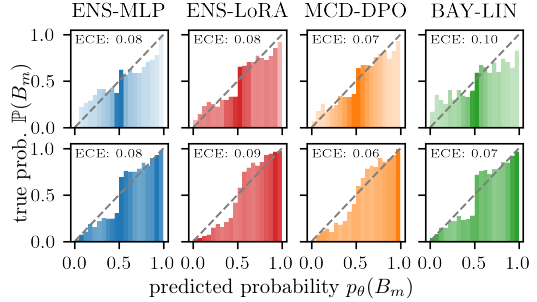}
        \caption{Calibration of predictions}
        \label{fig:results-calibration-ece}
    \end{subfigure}
    \begin{subfigure}{0.5\linewidth}
        \centering
        \includegraphics[width=\linewidth]{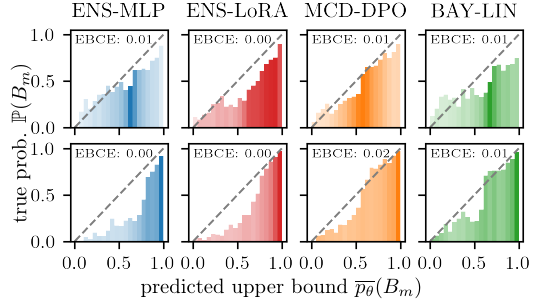}
        \caption{Calibration of (upper) bounds}
        \label{fig:results-calibration-ebce}
    \end{subfigure}
    \caption{
        Calibration diagrams for Qwen3-0.6B (top) and Qwen3-4B (bottom) trained on UltraFeedback and evaluated on RewardBench.
        The predictions are well-calibrated when they agree with the actual probability per bin (\ie{} on the diagonal), while the predicted upper bounds are well-calibrated when they consistently exceed the actual probability per bin (\ie{} below the diagonal).
        The calibration metrics are defined in \cref{eq:metrics-calibration-predictions,eq:metrics-calibration-bounds}.
        The color intensity of each bar is proportional to the bin size.
        As described in \cref{sec:uq-calibration}, the calibration diagrams for the upper and lower bounds are equivalent.
    }
    \label{fig:results-calibration}
\end{figure}

\paragraph{Insights into calibration}
We further analyze the calibration of our UQ methods, as defined in \cref{sec:uq-calibration}, for two representative initialization models, namely, Qwen3-0.6B and Qwen3-4B.
As shown in \cref{fig:results-calibration}, the different UQ methods provide similarly calibrated preference predictions and confidence bounds with ECE rates below $0.1$ and EBCE rates below $0.01$.
However, we observe that the distribution of predicted preference probabilities concentrates around $0.5$ for smaller models, as seen from the color intensities in \cref{fig:results-calibration-ece}, indicating lower certainty on average.
Similarly, we observe that smaller models tend to be slightly overconfident if certain, since the predicted preference probabilities for $> 0.5$ lie below the diagonal and for $< 0.5$ above the diagonal.
We discuss the unusual appearance of calibration diagrams in the context of preference classification in \cref{sec:theoretical_details-preference_classification}.


We provide detailed results and further discussion on both accuracy and calibration in \cref{sec:supplementary_results}.

\section{Discussion}\label{sec:discussion}

\paragraph{Conclusions}
Uncertainty quantification for reward models offers a promising direction for robust and sample-efficient RLHF, ultimately improving the safety and helpfulness of language models.
In this work, we introduced \emph{RewardUQ}, a unified framework to design and evaluate uncertainty-aware reward models, which complements prior work focusing on downstream applications exclusively.
The unified formalization of existing approaches, a novel metric balancing accuracy and uncertainty, and a common evaluation procedure enable a more systematic comparison of different methods.
Our results show that utilizing finetuned base models results in higher accuracy, but we find that the best-performing method remains instance dependent.
By releasing our framework as an open-source library, we aim to lower the barrier to uncertainty-aware alignment research and foster applications in active learning, safe alignment, and reward robustness.


\paragraph{Limitations} Our study focuses on the intrinsic evaluation of UQ methods rather than their performance in downstream reinforcement learning loops. This scope was chosen intentionally to disentangle the quality of uncertainty estimates from the confounding variables inherent in PPO or DPO fine-tuning. We posit that identifying robust UQ signals in isolation provides a more reliable foundation for researchers than costly end-to-end ablations. Furthermore, as detailed in \cref{sec:theoretical_details-ranking}, we acknowledge that our proposed ranking score (\cref{eq:metrics-accuracy-ranking}) entails specific trade-offs between calibration and discrimination that may not perfectly capture every nuance of downstream utility.


\paragraph{Future Work} To bolster the generalizability of our findings, future comparisons should expand to a broader range of algorithms, model families, datasets, and metrics. Additionally, while this work provides extensive empirical benchmarks, the theoretical mechanisms governing epistemic uncertainty in reward models for preference optimization remain under-explored. We hope this work serves as a launchpad for rigorous theoretical analysis and the development of next-generation applications in active preference learning and safety-constrained alignment.

\begin{impact}
The rigorous study of UQ for reward models offers promising improvements to various downstream applications, such as reducing the costs of data collection via active learning, enhancing the safety of LLMs via uncertainty-aware alignment, and making reward models more interpretable.
However, we acknowledge that highly accurate UQ methods could eventually serve as an additional attack vector targeting the uncertainties of LLMs.
While current methods do not yet pose a significant risk, we emphasize the importance of monitoring dual-use risks in the future.
\end{impact}

\begin{ack}
This work was supported as part of the Swiss AI initiative by a grant from the Swiss National Supercomputing Centre (CSCS) under project IDs \texttt{a10}, \texttt{a145}, and \texttt{infra01} on Alps.
Barna Pásztor was primarily supported by the ETH AI Center through an ETH AI Center doctoral fellowship, and Ido Hakimi primarily supported by the ETH AI Center through an ETH AI Center postdoctoral fellowship.
\end{ack}

\bibliography{references}
\bibliographystyle{setup/custom_natbib}


\newpage
\appendix

\renewcommand*\contentsname{Appendix}
\addtocontents{toc}{\protect\setcounter{tocdepth}{2}}
\tableofcontents

\section{Theoretical details}\label{sec:theoretical_details}

\subsection{Background on the uncertainty decomposition in preference classification}\label{sec:theoretical_details-uncertainty_decomposition}

Traditionally, the total uncertainty is decomposed into epistemic uncertainty, which describes the lack of knowledge in the model, and aleatoric uncertainty, which captures the irreducible randomness in the data \citep{hullermeier2021aleatoric, kendall2017what}.
The two commonly used decompositions of the total uncertainty in terms of the variance and in terms of the entropy have been explicitly derived by \citet{depeweg2018decomposition} and correspond to
\begin{alignat*}{2}
    \Var*{Y\mid X}
    &= \Var_{\theta}\brack*{\E_{Y}\brack*{Y\mid X;\theta}}
    &&+ \E_{\theta}\brack*{\Var_{Y}\brack*{Y\mid X;\theta}}
    \\ \Ent*{Y\mid X}
    &= I(Y;\theta)
    &&+ \E_{\theta}\brack*{\Ent*{Y\mid X,\theta}}
    ,
\end{alignat*}
with $X$ denoting the input, $Y$ the output and $\theta$ the parameters of the assumed underlying statistical model.
The first term describes the epistemic and the second term the aleatoric uncertainty.

In the context of preference classification described in \cref{sec:uq-problem}, $X$ denotes the sample $(x,y,y')$ and $Y$ the label $\indicator*{y\succ y'}$.
The common statistical model, which we adopt in this work, is given by
\begin{equation}\label{eq:statistical_model_assumption}
    y\succ y'\mid x,y,y' \sim \Ber*{\sigmoid*{\rmodel(x,y) - \rmodel(x,y')}}
\end{equation}
based on the Bradley-Terry model \citep{bradley1952rank}.
Importantly, this statistical model makes the following assumptions:
First, it assumes that the preference label contains Bernoulli noise, which leads to aleatoric uncertainty in the preference predictions.
Second, it assumes that the preference signal comes from a deterministic reward function through the Bradley-Terry model and, hence, is free of randomness.
Hence, under this statistical model assumption, reward models trained on preference data are free of aleatoric uncertainty and only contain epistemic uncertainty, as the aleatoric uncertainty is fully captured by the Bernoulli noise model.

Some work \citep{lou2025uncertaintyaware, yan2024rewardrobust} use the alternative statistical model assumption
\begin{align*}
    y\succ y'\mid R,R' &\sim \Ber*{\sigmoid*{R - R'}}
    \\ \text{with}\quad
    R\mid x,y &\sim \Normal*{\rmodel(x,y), s_\theta^2(x,y)}
    \\ R'\mid x,y' &\sim \Normal*{\rmodel(x,y'), s_\theta^2(x,y')}
\end{align*}
which assumes heteroscedastic Gaussian noise in the reward, leading to aleatoric uncertainty in the underlying reward models.
This assumption is equivalent to
\begin{align*}
    y\succ y'\mid x,y,y',\Delta\varepsilon &\sim \Ber*{\sigmoid*{\rmodel(x,y) - \rmodel(x,y') + \Delta\varepsilon}}
    \\ \text{with}\quad
    \Delta\varepsilon\mid x,y,y' &\sim \Normal*{0, s_\theta^2(x,y) + s_\theta^2(x,y')}
    .
\end{align*}
Intuitively, the Gaussian noise assumption in the reward smooths the sigmoid function with a Gaussian kernel with bandwidth $s_\theta^2(x,y) + s_\theta^2(x,y')$ as we marginalize over $\Delta\varepsilon$.
Hence, the smoothed sigmoid function converges towards a constant function at $0.5$ with increasing noise level, while the original sigmoid function is recovered with zero noise.

In summary, we adopt the more common statistical model assumption in \cref{eq:statistical_model_assumption} and assume the aleatoric uncertainty to be fully captured by the Bernoulli model, while the underlying reward model is free of aleatoric uncertainty.


\subsection{Background on the symmetry in preference classification}\label{sec:theoretical_details-preference_classification}

Pairwise preference classification is a special form of binary classification, where the goal is predict the label $\indicator{\set{y\succ y'}}\in\set{0,1}$ for a preference sample $(x,y,y')$.
Unlike standard binary classification, the label is defined through the antisymmetric relation $\succ$, which implies
\begin{equation*}
    \indicator{\set{y\succ y'}} = 1 - \indicator{\set{y'\succ y}}
    .
\end{equation*}
Hence, \emph{there is no distinction between positive and negative classes} in pairwise preference classification, since each sample $(x,y,y')$ is equivalent to its flipped counterpart $(x,y',y)$ with the class label inverted.
As a result, the predictive accuracy is fully characterized by the win rate defined in \cref{eq:metrics-accuracy-predictions}, which jointly describes the true positive (TP), true negative (TN), false positive (FP) and false negative (FN) rate as
\begin{equation*}
    \mathrm{win\ rate} = \mathrm{TP\ rate} = \mathrm{TN\ rate}
    \quad\text{and}\quad
    1 - \mathrm{win\ rate} = \mathrm{FP\ rate} = \mathrm{FN\ rate}
    .
\end{equation*}

The same antisymmetry extends to preference probabilities, yielding
\begin{equation*}
    p(y\succ y'\mid x,y,y') = 1 - p(y'\succ y\mid x,y',y)
    .
\end{equation*}
Similarly, an upper bound on a preference probability induces a corresponding lower bound for the flipped comparison and vice versa, \ie{}
\begin{equation*}
    \begin{aligned}
        \ub{p}(y\succ y'\mid x,y,y') &= 1 - \lb{p}(y'\succ y\mid x,y',y)
        \\ \lb{p}(y\succ y'\mid x,y,y') &= 1 - \ub{p}(y'\succ y\mid x,y',y)
        .
    \end{aligned}
\end{equation*}
In theory, this is irrelevant for computing the expected calibration errors defined in \cref{eq:metrics-calibration-predictions,eq:metrics-calibration-bounds}.
However, in practice, when approximating these errors via binning, it is essential to consider both $(x, y, y')$ and $(x, y', y)$, effectively doubling the evaluation set \citep{shen2024datacentric}.
This ensures that predictions for both $y\succ y'$ and $y'\succ y$ contribute to the empirical frequencies in the corresponding bins.
As a result, the calibration diagram for predictions is point-symmetric at $(0.5, 0.5)$,%
\footnote{This is why in a calibration diagram overconfidence appears as a flat line (\ie{} above the diagonal on $[0.0,0.5]$ and below the diagonal on $[0.5,1.0]$), while underconfidence forms a sigmoid-shaped curve.}
and the calibration diagrams for upper and lower bounds are equivalent, resulting in identical calibration errors for both.

\subsection{Background on the accuracy metrics}\label{sec:theoretical_details-accuracy}

Our extension of accuracy metrics to predictions under uncertainty in \cref{sec:uq-accuracy} can be generalized to standard binary classification metrics.
Specifically, categorizing predictions into confident and unconfident introduces an orthogonal dimension, resulting in a three-dimensional confusion tensor along the axes
\begin{equation*}
    \set{\mathbf{C}\mathrm{(onfident)}, \mathbf{U}\mathrm{(nconfident)}} \times \set{\mathbf{T}\mathrm{(rue)}, \mathbf{F}\mathrm{(alse)}} \times \set{\mathbf{P}\mathrm{(ositive)}, \mathbf{N}\mathrm{(egative)}},
\end{equation*}
where $\mathbf{P}$ and $\mathbf{N}$ denote the set of real positives and negatives, respectively.
For example, the confident true positive (CTP) rate is then defined as $\mathrm{CTP\ rate} = \frac{\abs{\mathbf{C} \cap \mathbf{T} \cap \mathbf{P}}}{\abs{\mathbf{P}}}$.

In preference classification, there is no distinction between positive and negative classes as described in \cref{sec:theoretical_details-preference_classification} and the confusion tensor collapses into the $2 \times 2$ matrix in \cref{eq:metrics-accuracy-bounds}, which should not be confused with the classical binary confusion matrix.
Accordingly, we normalize by the total number of samples instead of by the per-class counts.

\subsection{Background on the ranking score}\label{sec:theoretical_details-ranking}

Observe that all accuracy metrics introduced in \cref{eq:metrics-accuracy-predictions,eq:metrics-accuracy-bounds} can be expressed in terms of the four base counts
\begin{equation}\label{eq:ranking-base_counts}
    \begin{aligned}
        &
        && \eqnote{confident}
        &&
        && \eqnote{unconfident}
        \\ \eqnote{true} &
        \quad
        & CT &= \abs{\mathbf{C} \cap \mathbf{T}}
        && \uparrow
        \quad
        & UT &= \abs{\mathbf{U} \cap \mathbf{T}}
        && \searrow
        \\ \eqnote{false} &
        \quad
        & CF &= \abs{\mathbf{C} \cap \mathbf{F}}
        && \downarrow
        \quad
        & UF &= \abs{\mathbf{U} \cap \mathbf{F}}
        && \searrow
    \end{aligned}
\end{equation}
with $T = CT + UT$ and $F = UT + UF$.
Since the total number of samples is fixed to the size of the evaluation dataset $n = CT + UT + CF + UF$, there are only three degrees of freedom, capturing the overall accuracy of predictions under uncertainty.
A ranking strategy reduces these three degrees of freedom to a single score, effectively compressing two dimensions along which differently performing models are ranked equally, reflecting the inherent trade-offs made by the ranking strategy.

Recall our proposed ranking score in \cref{eq:metrics-accuracy-ranking}, which can be expressed in terms of these counts as
\begin{align*}
    \mathrm{RS}_\alpha
    &= \frac{\mathrm{CT\ rate}}{\mathrm{win\ rate} + \alpha \cdot (1 - \mathrm{win\ rate})} - \frac{\mathrm{CF\ rate}}{(1 - \mathrm{win\ rate}) + \alpha \cdot \mathrm{win\ rate}}
    \\ &= \frac{CT}{T + \alpha \cdot F} - \frac{CF}{F + \alpha \cdot T}
    ,
\end{align*}
where $\alpha\in[0,1]$ balances the inherent trade-offs between the three degrees of freedom.
The general idea is to encourage confident true predictions and penalize confident false predictions.
Depending on the choice of $\alpha$, the score normalizes the number of confident predictions differently and puts a different focus on confidence and accuracy.
In the following, we first discuss the two edge cases $\alpha=0$ and $\alpha=1$ and then how our ranking score formulation unifies both cases.

\begin{figure}
    \centering
    \begin{subfigure}{\linewidth}
        \centering
        \includegraphics[width=\linewidth]{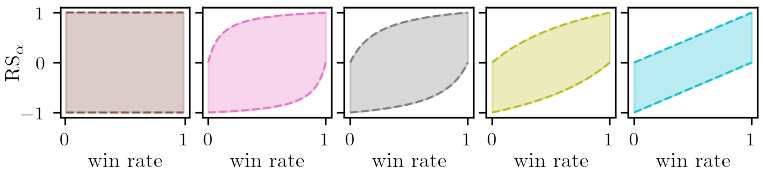}
        \caption{Ranking score ranges}
        \label{fig:ranking-range}
    \end{subfigure}
    \par\medskip
    \begin{subfigure}{0.4\linewidth}
        \centering
        \includegraphics[width=\linewidth]{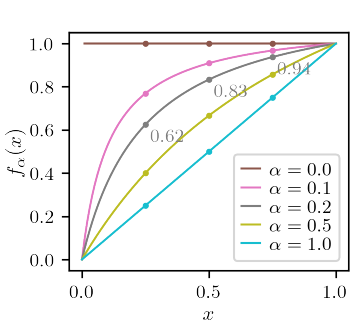}
        \vspace*{-3.5ex}
        \caption{Ranking score weights}
        \label{fig:ranking-factors}
    \end{subfigure}%
    \caption{
        Background on our ranking score for different $\alpha$.
        While the range is invariant of the win rate for $\alpha=0$, it has a linear dependence for $\alpha=1$ as shown in \cref{fig:ranking-range}.
        The inherent trade-off underlying the choice of $\alpha$ is shown in \cref{fig:ranking-factors}, which visualizes the weights in our ranking score in \cref{eq:ranking-unified}.
        For example, with $\alpha=0.2$, when the win rate increases from $0.6$ to $0.8$, the confidence among true predictions is upweighted from $0.88$ to $0.95$ by a factor of $\approx 1.08$, while the confidence among false predictions is downweighted from $0.77$ to $0.56$ by a factor of $\approx 0.73$.
    }
    \label{fig:ranking}
\end{figure}

\paragraph{Focus on confidence ($\mathrm{RS}_0$)}
When $\alpha=0$, the ranking score corresponds to
\begin{equation}\label{eq:ranking-0}
    \mathrm{RS}_0
    = \frac{\mathrm{CT\ rate}}{\mathrm{win\ rate}} - \frac{\mathrm{CF\ rate}}{1 - \mathrm{win\ rate}}
    = \frac{CT}{T} - \frac{CF}{F}
    \in [-1,1]
    .
\end{equation}
This ranking score considers the relative rate of confidence among true and false predictions.
In other words, it looks at the proportions of confident predictions conditioned on the correctness of the predictions.
Intuitively, it encourages confidence among true predictions and penalizes confidence among false predictions, while it does not take into account the overall ratio between true and false predictions.
In particular, observe that the range of this ranking score is invariant of the win rate as shown in \cref{fig:ranking-range}.
Overall, this ranking score focuses on the confidence of predictions.


\textit{Remark:} This ranking score is conceptually related to Youden's index \citep{youden1950index}, a summary statistic in binary classification, which is defined as
$J = \frac{TP}{TP + FN} - \frac{FP}{TN + FP} = \frac{\abs{\mathbf{T} \cap \mathbf{P}}}{\abs{\mathbf{P}}} - \frac{\abs{\mathbf{F} \cap \mathbf{N}}}{\abs{\mathbf{N}}}$.

\paragraph{Focus on accuracy ($\mathrm{RS}_1$)}
When $\alpha=1$, the ranking score corresponds to
\begin{equation}\label{eq:ranking-1}
    \mathrm{RS}_1
    = \mathrm{CT\ rate} - \mathrm{CF\ rate}
    = \frac{CT}{n} - \frac{CF}{n}
    \in [\mathrm{win\ rate} - 1, \mathrm{win\ rate}]
    .
\end{equation}
This ranking score considers the absolute rate of confidence among all samples.
In other words, it looks at the joint proportions of confidence and correctness.
Intuitively, it encourages confident correctness and penalizes confident incorrectness among all samples, while it does not take into account the number of uncertain true and false predictions.
Observe that the range of this ranking score is determined based on this win rate as shown in \cref{fig:ranking-range}.
Overall, this ranking score focuses on the confidence and accuracy of predictions.

\paragraph{Unified formulation ($\mathrm{RS}_\alpha$)}
The issue of $\mathrm{RS}_0$ is that it focuses too much on the confidence and cannot distinguish between models with completely different win rates, as long as the relative proportions of confidence are the same.
The issue of $\mathrm{RS}_1$ is that it focuses too much on the accuracy and, when setting the uncertainty globally to zero, it simplifies to $\mathrm{RS}_1 = 2 \cdot \mathrm{win\ rate} - 1$.
Hence, we introduce $\alpha\in[0,1]$ to balance the trade-off between both ends.
To better understand the effect of $\alpha$, we can rewrite our ranking score formulation into
\begin{equation}\label{eq:ranking-unified}
    \mathrm{RS}_\alpha
    = f_\alpha(\mathrm{win\ rate}) \cdot \frac{\mathrm{CT\ rate}}{\mathrm{win\ rate}} - f_\alpha(1- \mathrm{win\ rate}) \cdot \frac{\mathrm{CF\ rate}}{1 - \mathrm{win\ rate}}
\end{equation}
which considers the relative rate of confidence, but each weighted by some factor $f_\alpha(x) = \frac{x}{x + \alpha \cdot (1 - x)}$ depending on the win rate.
Intuitively, a higher win rate puts more weight on the bonus caused by confidence among true predictions, while a lower win rate puts more weight on the penalty caused by confidence among false predictions.
We visualize $f_\alpha(x)$ for different choices of $\alpha$ in \cref{fig:ranking-factors}.
In this work, we choose $\alpha=0.2$ as it balances well the trade-off between confidence and accuracy.

\subsubsection{Invariances of $\mathrm{RS}_0$}\label{sec:theoretical_details-ranking-invariances_rs0}

Recall \cref{eq:ranking-0}, which can be expressed in terms of the base counts from \cref{eq:ranking-base_counts} as
\begin{equation*}
    \mathrm{RS}_0 = \frac{CT}{CT + UT} - \frac{CF}{CF + UF}
    .
\end{equation*}
This ``difference of two ratios'' introduces two invariances, as analyzed in the following.

\paragraph{Invariance 1 (by normalization)}
One degree of freedom is lost due to normalization in the ratios. 
Hence, $\mathrm{RS}_0$ is invariant to changes in the numerator and denominator, as long as each ratio is preserved, \eg{}
\begin{center}
    \begin{tabular}{cccc|c}
        \toprule
        $CT$ & $UT$ & $CF$ & $UF$ & $\mathrm{RS}_0$ \\
        \midrule
        $40$ & $60$ & $2$ & $8$ & $0.2 \mathcolor{gray}{= 0.4 - 0.2}$ \\
        $42$ & $63$ & $1$ & $4$ & $0.2 \mathcolor{gray}{= 0.4 - 0.2}$ \\
        \bottomrule
    \end{tabular}
\end{center}
Intuitively, two models with different numbers of true and false predictions (\ie{} different win rates) are ranked equally if they share the same proportion of confident predictions among true and false predictions.
Formally, we can scale the numerators and denominators in each ratio equally
\begin{equation*}
    \begin{aligned}
        &
        && \eqnote{confident}
        && \eqnote{unconfident}
        \\ \eqnote{true} &
        \quad
        & CT &\to \paren*{1+\frac{\delta}{T}} CT
        \quad
        & UT &\to \paren*{1+\frac{\delta}{T}} UT
        \\ \eqnote{false} &
        \quad
        & CF &\to \paren*{1-\frac{\delta}{F}} CF
        \quad
        & UF &\to \paren*{1-\frac{\delta}{F}} UF
    \end{aligned}
\end{equation*}
based on some $\delta$.%
\footnote{For simplicity, we omit that $\delta$ must ensure non-negative integer counts.}
The resulting ranking score
\begin{equation*}
    \begin{aligned}
        \mathrm{RS}_0(\delta)
        &= \frac{\paren*{1+\frac{\delta}{T}} CT}{\paren*{1+\frac{\delta}{T}} CT + \paren*{1+\frac{\delta}{T}} UT} - \frac{\paren*{1-\frac{\delta}{F}} CF}{\paren*{1-\frac{\delta}{F}} CF + \paren*{1-\frac{\delta}{F}} UF}
        \\ &= \frac{CT}{CT + UT} - \frac{CF}{CF + UF}
        \\ &= \text{const.}
    \end{aligned}
\end{equation*}
is independent of $\delta$ and thus constant.

\paragraph{Invariance 2 (by taking the difference)}
The other degree of freedom is eliminated by taking the difference between the two terms.
Accordingly, $\mathrm{RS}_0$ is invariant to changes in both terms, provided their difference stays constant, \eg{}
\begin{center}
    \begin{tabular}{cccc|c}
        \toprule
        $CT$ & $UT$ & $CF$ & $UF$ & $\mathrm{RS}_0$ \\
        \midrule
        $40$ & $60$ & $2$ & $8$ & $0.2 \mathcolor{gray}{= 0.4 - 0.2}$ \\
        $70$ & $30$ & $5$ & $5$ & $0.2 \mathcolor{gray}{= 0.7 - 0.5}$ \\
        \bottomrule
    \end{tabular}
\end{center}
Intuitively, two models with different degrees of confidence are ranked equally if the true and false predictions are affected by the confidence level similarly.
Formally, we can increase or decrease the number of confident predictions for both true and false predictions
\begin{equation*}
    \begin{aligned}
        &
        && \eqnote{confident}
        && \eqnote{unconfident}
        \\ \eqnote{true} &
        \quad
        & CT &\to CT + \delta \cdot T
        \quad
        & UT &\to UT - \delta \cdot T
        \\ \eqnote{false} &
        \quad
        & CF &\to CF + \delta \cdot F
        \quad
        & UF &\to UF - \delta \cdot F
    \end{aligned}
\end{equation*}
based on some $\delta$.
The resulting ranking score
\begin{equation*}
    \begin{aligned}
        \mathrm{RS}_0(\delta)
        &= \frac{CT + \delta \cdot T}{(CT + \delta \cdot T) + (UT - \delta \cdot T)} - \frac{CF + \delta \cdot F}{(CF + \delta \cdot F) + (UF - \delta \cdot F)}
        \\ &= \frac{CT}{CT + UT} - \frac{CF}{CF + UF}
        \\ &= \text{const.}
    \end{aligned}
\end{equation*}
remains independent of $\delta$, confirming the invariance.

\subsubsection{Invariances of $\mathrm{RS}_1$}\label{sec:theoretical_details-ranking-invariances_rs1}

Recall \cref{eq:ranking-1}, which can be expressed in terms of the base counts from \cref{eq:ranking-base_counts} as
\begin{equation*}
    \mathrm{RS}_1 = \frac{CT}{CT + UT + UF + CF} - \frac{CF}{CT + UT + UF + CF}
    .
\end{equation*}
Note that both denominators are constant, as they correspond to the number of evaluation samples.

\paragraph{Invariance 1 (by indistinction between unconfident predictions)}
One degree of freedom is lost by not considering the base counts $UT$ and $UF$ separately.
Hence, $\mathrm{RS}_1$ is invariant to changes in both counts as long as their sum is preserved, \eg{}
\begin{center}
    \begin{tabular}{cccc|c}
        \toprule
        $CT$ & $UT$ & $CF$ & $UF$ & $\mathrm{RS}_1$ \\
        \midrule
        $40$ & $60$ & $2$ & $8$  & $0.35 \mathcolor{gray}{\approx 40/110 - 2/110}$ \\
        $40$ & $8$  & $2$ & $60$ & $0.35 \mathcolor{gray}{\approx 40/110 - 2/110}$ \\
        \bottomrule
    \end{tabular}
\end{center}
Intuitively, the ranking score does not distinguish between unconfident true and false predictions.
Formally, we can change the base counts of unconfident predictions
\begin{equation*}
    \begin{aligned}
        &
        && \eqnote{confident}
        && \eqnote{unconfident}
        \\ \eqnote{true} &
        \quad
        & CT &\to CT
        \quad
        & UT &\to UT + \delta
        \\ \eqnote{false} &
        \quad
        & CF &\to CF
        \quad
        & UF &\to UF - \delta
    \end{aligned}
\end{equation*}
based on some $\delta$.
The resulting ranking score
\begin{equation*}
    \begin{aligned}
        \mathrm{RS}_1(\delta)
        &= \frac{CT}{CT + \paren*{UT + \delta} + \paren*{UF - \delta} + CF} - \frac{CF}{CT + \paren*{UT + \delta} + \paren*{UF - \delta} + CF}
        \\ &= \frac{CT}{CT + UT + UF + CF} - \frac{CF}{CT + UT + UF + CF}
        \\ &= \text{const.}
    \end{aligned}
\end{equation*}
is independent of $\delta$ and thus constant.

\paragraph{Invariance 2 (by taking the difference)}
The other degree of freedom is eliminated by taking the difference between the two terms.
Accordingly, $\mathrm{RS}_1$ is invariant to changes in both terms, provided their difference stays constant, \eg{}
\begin{center}
    \begin{tabular}{cccc|c}
        \toprule
        $CT$ & $UT$ & $CF$ & $UF$ & $\mathrm{RS}_1$ \\
        \midrule
        $40$ & $60$ & $2$ & $8$ & $0.35 \mathcolor{gray}{\approx 40/110 - 2/110}$ \\
        $48$ & $52$ & $10$ & $0$ & $0.35 \mathcolor{gray}{\approx 48/110 - 10/110}$ \\
        \bottomrule
    \end{tabular}
\end{center}
Intuitively, two models with different degrees of confidence are ranked equally if the true and false predictions are affected by the confidence level similarly.
Formally, we can increase or decrease the number of confident predictions for both true and false predictions
\begin{equation*}
    \begin{aligned}
        &
        && \eqnote{confident}
        && \eqnote{unconfident}
        \\ \eqnote{true} &
        \quad
        & CT &\to CT + \delta
        \quad
        & UT &\to UT - \delta
        \\ \eqnote{false} &
        \quad
        & CF &\to CF + \delta
        \quad
        & UF &\to UF - \delta
    \end{aligned}
\end{equation*}
based on some $\delta$.
The resulting ranking score
\begin{equation*}
    \begin{aligned}
        \mathrm{RS}_1(\delta)
        &= \begin{multlined}[t]
            \frac{\paren*{CT + \delta}}{\paren*{CT + \delta} + \paren*{UT - \delta} + \paren*{UF - \delta} + \paren*{CF + \delta}}
            \\ - \frac{\paren*{CF + \delta}}{\paren*{CT + \delta} + \paren*{UT - \delta} + \paren*{UF - \delta} + \paren*{CF + \delta}}
        \end{multlined}
        \\ &= \frac{CT}{CT + UT + UF + CF} - \frac{CF}{CT + UT + UF + CF}
        \\ &= \text{const.}
    \end{aligned}
\end{equation*}
remains independent of $\delta$, confirming the invariance.

\subsection{Background on the calibration metrics}\label{sec:theoretical_details-calibration}
For readability, we use $Y^{\succ} \in \set{0,1}$ to denote the event $y \succ y'$ for given $x,y$ and $y'$.

\paragraph{Calibration of predictions}
The predicted preference probabilities are well-calibrated if they match the true preference probabilities%
, \ie{}
\begin{equation*}
    \ptrue(Y^{\succ} \mid \pmodel(Y^{\succ}) = p) = p
\end{equation*}
for all \(p \in [0,1]\), following \citet{guo2017calibration}.
The \emph{expected calibration error (ECE)} is defined as
\begin{equation*}
    \mathrm{ECE} = \E_p\brack*{\abs*{\ptrue(Y^{\succ} \mid \pmodel(Y^{\succ}) = p) - p}}
    ,
\end{equation*}
which penalizes over- and underestimations of the true preference probabilities.
Since the true probabilities are unknown in practice, we measure the deviation from the empirical probabilities.
Specifically, the predicted probabilities are grouped into $M$ bins $\set{B_m}_{m=1}^M$ and we compute
\begin{equation*}
     \begin{alignedat}{3}
        \mathrm{ECE} &\approx \sum_{m=1}^{M} \frac{\abs{B_m}}{n} \abs*{\ptrue(B_m) - \pmodel(B_m)} && \quad\downarrow
    \end{alignedat}
\end{equation*}
with empirical probability $\ptrue(B_m)$ and average predicted probability $\pmodel(B_m)$ in each bin $B_m$ \citep{guo2017calibration, pavlovic2025understanding}, as stated in \cref{eq:metrics-calibration-predictions}.

\paragraph{Calibration of bounds}
The predicted preference probability bounds are well-calibrated if they are not violated by the true preference probabilities, \ie{}
\begin{equation*}
    \begin{aligned}
        \ptrue(Y^{\succ} \mid \lb{\pmodel}(Y^{\succ}) = \lb{p}) &\geq \lb{p} \\
        \ptrue(Y^{\succ} \mid \ub{\pmodel}(Y^{\succ}) = \ub{p}) &\leq \ub{p}
    \end{aligned}
\end{equation*}
for all $\lb{p},\ub{p}\in[0,1]$.
We introduce the \emph{expected lower calibration error (ELCE)} and \emph{expected upper calibration error (EUCE)} as follows
\begin{equation*}
    \begin{aligned}
        \mathrm{ELCE} &= \E_{\lb{p}}\brack*{\max\paren*{\lb{p} - \ptrue\paren*{Y^{\succ} \mid \lb{\pmodel}(Y^{\succ}} = \lb{p}), 0}} \\
        \mathrm{EUCE} &= \E_{\ub{p}}\brack*{\max\paren*{\ptrue\paren*{Y^{\succ} \mid \ub{\pmodel}(Y^{\succ}) = \ub{p}} - \ub{p}, 0}}
    \end{aligned}
\end{equation*}
with ELCE penalizing lower bounds that overestimate the true preference probability and EUCE penalizing upper bounds that underestimate the true preference probability.
In practice, analogous to ECE, we group the lower and upper bounds separately into $M$ bins $\set{B_m}_{m=1}^M$ and compute
\begin{equation*}
    \begin{alignedat}{3}
        \mathrm{ELCE} &\approx \sum_{m=1}^{M} \frac{\abs{B_m}}{n} \max\paren*{\lb{\pmodel}(B_m) - \ptrue(B_m), 0} && \quad\downarrow
        \\ \mathrm{EUCE} &\approx \sum_{m=1}^{M} \frac{\abs{B_m}}{n} \max\paren*{\ptrue(B_m) - \ub{\pmodel}(B_m), 0} && \quad\downarrow
    \end{alignedat}
\end{equation*}
with $\lb{\pmodel}(B_m)$ and $\ub{\pmodel}(B_m)$ denoting the average predicted lower and upper bounds in the corresponding bin $B_m$, as stated in \cref{eq:metrics-calibration-bounds}.



\section{Experimental details}\label{sec:experimental_details}

\subsection{Technical setup}

All models were trained on a single node equipped with four NVIDIA GH200 GPUs, providing a total of 378GB of VRAM.
Our implementation is built on top of \emph{Transformers} \citep{wolf2020transformers} and \emph{TRL} \citep{vonwerra2025trl} by HuggingFace, with multi-GPU management handled by \emph{Accelerate} \citep{gugger2022accelerate}.
We use data parallelism for models that fit on a single GPU, and model and tensor parallelism for larger models.

\subsection{Hyperparameters}\label{sec:experimental_details-hyperparameters}

We used the AdamW optimizer \citep{loshchilov2018decoupled} (with a weight decay of 0, $\beta_1=0.9$, $\beta_2=0.999$, and $\epsilon=10^{-8}$) and an effective batch size of 64 via gradient accumulation for all experiments, except for the LoRA adapter ensemble, which used a batch size of 16.
We trained our models using a single epoch with a cosine learning rate scheduler and a warm-up phase of 5\% of the total number of steps.
The reward uncertainty bounds in \cref{eq:reward-bounds} are constructed using $\beta = 2$ for the MLP head ensemble and DPO-based MC dropout and $\beta=0.5$ for the Bayesian linear head. The best hyperparameters for each base model are given in \cref{tab:hyperparameters}.

\paragraph{MLP head ensemble}
We use an ensemble of $K=20$ MLP heads.
Each head is a two-layer network with 128 nodes per layer and ReLU activation functions.
They are initialized using a Xavier uniform distribution with a gain of one. 
We conducted a grid search over the learning rate $\eta\in\set{10^{-5}, 10^{-4}, 10^{-3}}$ and regularization parameters $\lambda\in\set{0.0, 0.1, 1.0}$ and $\gamma\in\set{0.0, 0.01, 0.1}$ from \cref{eq:models-ens_mlp-loss}.

\paragraph{LoRA adapter ensemble}
We use an ensemble of $K=8$ LoRA adapter with rank $r_\text{LoRA} = 16$ and scaling factor $\alpha_\text{LoRA} = 32$.
We conducted a grid search over the learning rate $\eta\in\set{10^{-5}, 10^{-4}, 10^{-3}}$ and regularization parameters $\lambda\in\set{0.001, 0.01, 0.1}$ and choose $\gamma=0.01$.

\paragraph{DPO-based MC dropout}
The ensemble is formed by sampling $K = 20$ dropout masks at inference time.
We conducted a grid search over the learning rate $\eta\in\set{10^{-7}, 10^{-6}, 10^{-5}}$, DPO regularization parameter $\lambda\in\set{0.01, 0.05, 0.1}$ and dropout rate $p_{\text{dropout}}\in\set{0.01, 0.05, 0.1}$ from \cref{eq:models-dpo_mcd-loss}.

\paragraph{Bayesian linear head}
We conducted a grid search over the learning rate $\eta\in\set{10^{-3}, 10^{-2}, 10^{-1}}$ and $\ell_2$-regularization parameter $\lambda\in\set{10^{-3}, 10^{-2}, 10^{-1}}$ from \cref{eq:models-bay_lin-loss}.


\begin{sidewaystable}
    \caption{Best hyperparameters found according to our evaluation procedure described in \cref{sec:experiments-setup}.}
    \label{tab:hyperparameters}
    \centering
    \begin{tabular}{rllllllllllllll}
        \toprule
        \multirow{3}{*}{base model} & \multicolumn{3}{c}{\textbf{ENS-MLP}} && \multicolumn{3}{c}{\textbf{ENS-LoRA}} && \multicolumn{3}{c}{\textbf{MCD-DPO}} && \multicolumn{2}{c}{\textbf{BAY-LIN}} \\
        \cmidrule{2-4} \cmidrule{6-8} \cmidrule{10-11}
        & \multicolumn{1}{c}{$\eta$} & \multicolumn{1}{c}{$\lambda$} & \multicolumn{1}{c}{$\gamma$} && \multicolumn{1}{c}{$\eta$} & \multicolumn{1}{c}{$\lambda$} & \multicolumn{1}{c}{$\gamma$} && \multicolumn{1}{c}{$\eta$} & \multicolumn{1}{c}{$\lambda$} & \multicolumn{1}{c}{$p_{\text{dropout}}$} && \multicolumn{1}{c}{$\eta$} & \multicolumn{1}{c}{$\lambda$} \\
        \midrule
        Qwen3-0.6B         & 0.001 & 0   & 0.01 && 0.0001 & 0.01 & 0.01 && 0.00001 & 0.01 & 0.05 && 0.01 & 0.01 \\
        Qwen3-1.7B         & 0.001 & 0   & 0.01 && 0.0001 & 0.01
 & 0.01 && 0.00001 & 0.01 & 0.1  && 0.001 & 0.01 \\
        Qwen3-4B           & 0.001 & 0.1 & 0    && 0.0001 & 0.001
 & 0.01 && 0.00001 & 0.01 & 0.1  && 0.001 & 0.001 \\
        Qwen3-8B           & 0.001 & 0.1 & 0    && & & && 0.001 & 0.01 \\
        Qwen3-14B          & 0.001 & 1   & 0    && & & && 0.001 & 0.01 \\
        Qwen3-32B          & 0.001 & 1   & 0.01 && & & && 0.001 & 0.1 \\
        \addlinespace[2pt]
        Skywork-Qwen3-0.6B & 0.001 & 0.1 & 0.01 && 0.0001 & 0.01
 & 0.01 && 0.00001 & 0.01 & 0.1  && 0.1 & 0.001 \\
        Skywork-Qwen3-1.7B & 0.001 & 1   & 0.01 && 0.0001 & 0.01 & 0.01 && 0.00001 & 0.01 & 0.1  && 0.1 & 0.01 \\
        Skywork-Qwen3-4B   & 0.001 & 0.1 & 0.01 && 0.0001 & 0.001 & 0.01 && 0.00001 & 0.01 & 0.05 && 0.1 & 0.1 \\
        Skywork-Qwen3-8B   & 0.001 & 1   & 0.01 && & & && 0.1 & 0.1 \\
        \bottomrule
    \end{tabular}
\end{sidewaystable}

\subsection{Dataset preprocessing}

For all datasets, we remove preference samples if the total sequence length of the prompt and one of the completions exceeds 2048 tokens to avoid additional evaluation noise due to the truncation of the prompt and completions.

\subsection{Evaluation on RewardBench}

RewardBench \citep{lambert2024rewardbench} consists of the four main categories ``Chat'', ``Chat Hard'', ``Safety'', and ``Reasoning'', each with weighted subcategories.
Following the standard procedure, we first compute our metrics from \cref{sec:uq-accuracy,sec:uq-calibration} (excluding the ranking score) within each category using the subcategory weights, then average them across categories, and finally derive the ranking score from these averages.

\FloatBarrier
\section{Supplementary results}\label{sec:supplementary_results}

\begin{figure}[ht]
    \centering
    \begin{subfigure}{\linewidth}
        \centering
        \includegraphics[width=\linewidth]{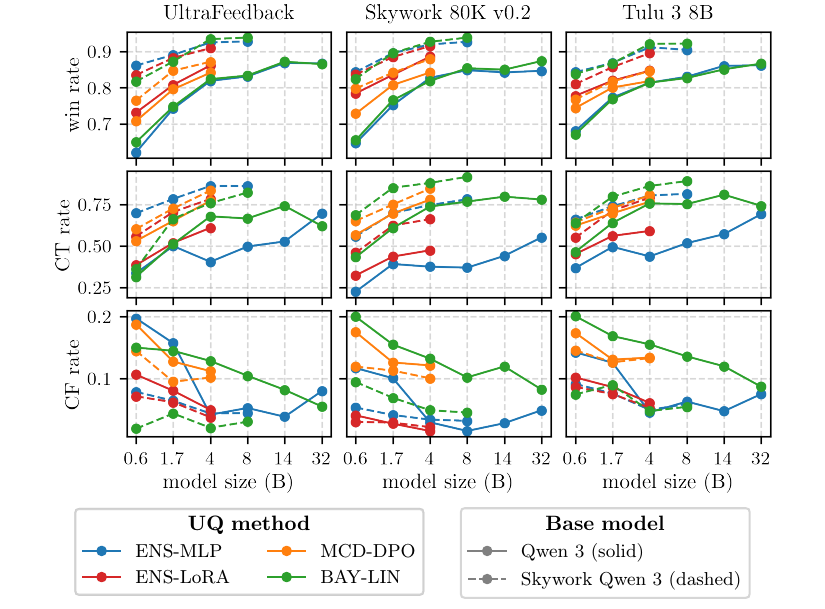}
        \caption{Accuracy metrics}
        \label{fig:supplementary_results-metrics-accuracy}
    \end{subfigure}
    \par\medskip
    \begin{subfigure}{\linewidth}
        \centering
        \includegraphics[width=\linewidth]{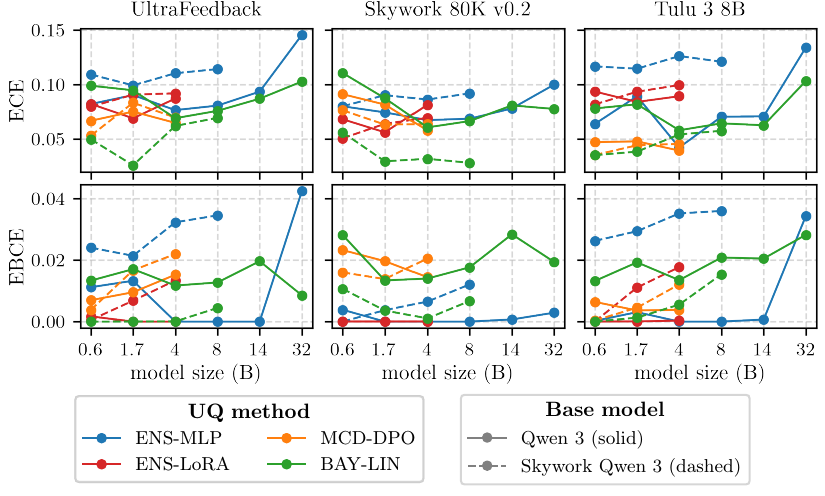}
        \caption{Calibration metrics}
        \label{fig:supplementary_results-metrics-calibration}
    \end{subfigure}
    \caption{
        Our base metrics on RewardBench across different UQ methods, training datasets, pretrained and finetuned models, and model sizes.
        The accuracy metrics are defined in \cref{eq:metrics-accuracy-predictions,eq:metrics-accuracy-bounds} and the calibration metrics in \cref{eq:metrics-calibration-predictions,eq:metrics-calibration-bounds}.
    }
    \label{fig:supplementary_results-metrics}
\end{figure}

\cref{fig:supplementary_results-metrics} provides detailed results supplementing the discussion in \cref{sec:experiments-results}.
\cref{fig:supplementary_results-metrics-accuracy} provides a granular decomposition of the ranking score trends, explicitly isolating the contributions of win rate, Confident True rate, and Confident False rate.
Validating the importance of initialization, methods utilizing the task-aligned Skywork base model consistently outperform the generic Qwen 3 base models across both win rate and CT rate. This advantage is particularly pronounced for fixed-head methods like BAY-LIN, which rely heavily on high-quality embeddings to produce confident, correct predictions. Furthermore, the diminishing returns in ranking scores observed in \cref{sec:experiments-results} are elucidated by the behavior of the CF rate relative to other metrics. While win rates generally improve or plateau with model size, the CF rate decreases at a lower rate than the CT rate increases.

Regarding calibration, \cref{fig:supplementary_results-metrics-calibration} supports the general observation that most UQ methods maintain reasonable calibration, with Expected Calibration Error (ECE) and Expected Bound Calibration Error (EBCE) typically remaining below 0.10 and 0.04, respectively.
However, the breakdown reveals specific instabilities that impact the aggregate performance; for example, ENS-MLP trained on the generic Qwen 3 base exhibits a sharp spike in EBCE at the 32B scale. This degradation in bound calibration aligns with the underperformance of ENS-MLP in the main ranking results for that configuration, suggesting that larger model sizes can lead to overfitting instabilities.

Overall, these supplementary metrics confirm that the superior ranking of task-aligned models is driven not just by higher accuracy, but by a more favorable balance of maximizing confident true predictions while suppressing confident errors.

\end{document}